\documentclass[sigconf]{acmart}

\usepackage{multirow}
\usepackage{booktabs}
\usepackage{cleveref}
\usepackage{subcaption}
\usepackage{xcolor,colortbl}
\usepackage{tikz}
\usepackage{booktabs}
\usepackage{multicol}
\usepackage{multirow}
\usepackage{todonotes}
\usepackage{mathtools}
\usepackage{balance}

\DeclarePairedDelimiter\floor{\lfloor}{\rfloor}

\usepackage{pifont}
\newcommand{\cmark}{\ding{51}}%
\newcommand{\xmark}{\ding{55}}%

\usepackage{xcolor}
\usetikzlibrary{positioning, shapes.geometric,decorations.pathreplacing,calc}
\definecolor{fc}{HTML}{1E90FF}
\definecolor{h}{HTML}{228B22}
\definecolor{bias}{HTML}{87CEFA}
\definecolor{noise}{HTML}{8B008B}
\definecolor{conv}{HTML}{FFA500}
\definecolor{pool}{HTML}{B22222}
\definecolor{up}{HTML}{B22222}
\definecolor{view}{HTML}{FFFFFF}
\definecolor{bn}{HTML}{FFD700}
\tikzset{fc/.style={black,draw=black,fill=fc,rectangle,minimum height=0.5cm}}
\tikzset{h/.style={black,draw=black,fill=h,rectangle,minimum height=0.5cm}}
\tikzset{bias/.style={black,draw=black,fill=bias,rectangle,minimum height=1cm}}
\tikzset{noise/.style={black,draw=black,fill=noise,rectangle,minimum height=1cm}}
\tikzset{conv/.style={black,draw=black,fill=conv,rectangle,minimum height=0.5cm}}
\tikzset{pool/.style={black,draw=black,fill=pool,rectangle,minimum height=1cm}}
\tikzset{up/.style={black,draw=black,fill=up,rectangle,minimum height=1cm}}
\tikzset{view/.style={black,draw=black,fill=view,rectangle,minimum height=1cm}}
\tikzset{bn/.style={black,draw=black,fill=bn,rectangle,minimum height=1cm}}

\definecolor{Gray}{gray}{0.85}
\definecolor{LightGreen}{rgb}{0.5,0.9,0.5}
\definecolor{LightCyan}{rgb}{0.5,0.9,0.9}
\definecolor{LightRed}{rgb}{0.9,0.5,0.5}

\AtBeginDocument{%
  }

\setcopyright{acmlicensed}
\copyrightyear{2018}
\acmYear{2018}
\acmDOI{XXXXXXX.XXXXXXX}

\acmConference[Conference acronym 'XX]{Make sure to enter the correct
  conference title from your rights confirmation emai}{June 03--05,
  2018}{Woodstock, NY}
\acmISBN{978-1-4503-XXXX-X/18/06}

\begin{document}

\title{Tokenization of Gaze Data}

\author{Tim Rolff}
\email{tim.rolff@uni-hamburg.de}
\orcid{0000-0001-9038-3196}
\affiliation{%
  \institution{University of Hamburg}
  \city{Hamburg}
  \country{Germany}
}
\author{Jurik Karimian}
\email{jurik.karimian@studium.uni-hamburg.de}
\affiliation{%
  \institution{University of Hamburg}
  \city{Hamburg}
  \country{Germany}
}

\author{Niklas Hypki}
\email{niklas.hypki@uni-muenster.de}
\affiliation{%
  \institution{University of Münster}
  \city{Münster}
  \country{Germany}
}

\author{Susanne Schmidt}
\email{susanne.schmidt@uni-hamburg.de}
\orcid{0000-0002-8162-7694}
\affiliation{%
  \institution{Canterbury University}
  \city{Christchurch}
  \country{New Zealand}
}

\author{Markus Lappe}
\email{mlappe@uni-muenster.de}
\orcid{0000-0001-8814-7098}
\affiliation{%
  \institution{University of Münster}
  \city{Muenster}
  \country{Germany}
}

\author{Frank Steinicke}
\email{frank.steinicke@uni-hamburg.de}
\orcid{0000-0001-9879-7414}
\affiliation{%
  \institution{University of Hamburg}
  \city{Hamburg}
  \country{Germany}
}

\newcommand{\best}[1]{\underline{\textbf{#1}}}

\renewcommand{\shortauthors}{Rolff et al.}


\begin{abstract}
A considerable part of the performance of today's large language models (LLM's) and multimodal large language models (MLLM's) depends on their tokenization strategies. While tokenizers are extensively researched for textual and visual input, there is no research on tokenization strategies for gaze data due to its nature. However, a corresponding tokenization strategy would allow using the vision capabilities of pre-trained MLLM's for gaze data, for example, through fine-tuning.\\

In this paper, we aim to close this research gap by analyzing five different tokenizers for gaze data on three different datasets for the forecasting and generation of gaze data through LLMs (cf.~\cref{fig:teaser}). We evaluate the tokenizers regarding their reconstruction and compression abilities. Further, we train an LLM for each tokenization strategy, measuring its generative and predictive performance. Overall, we found that a quantile tokenizer outperforms all others in predicting the gaze positions and k-means is best when predicting gaze velocities.
%
\end{abstract}


\begin{teaserfigure}
  \centering
  \includegraphics[width=0.92\textwidth]{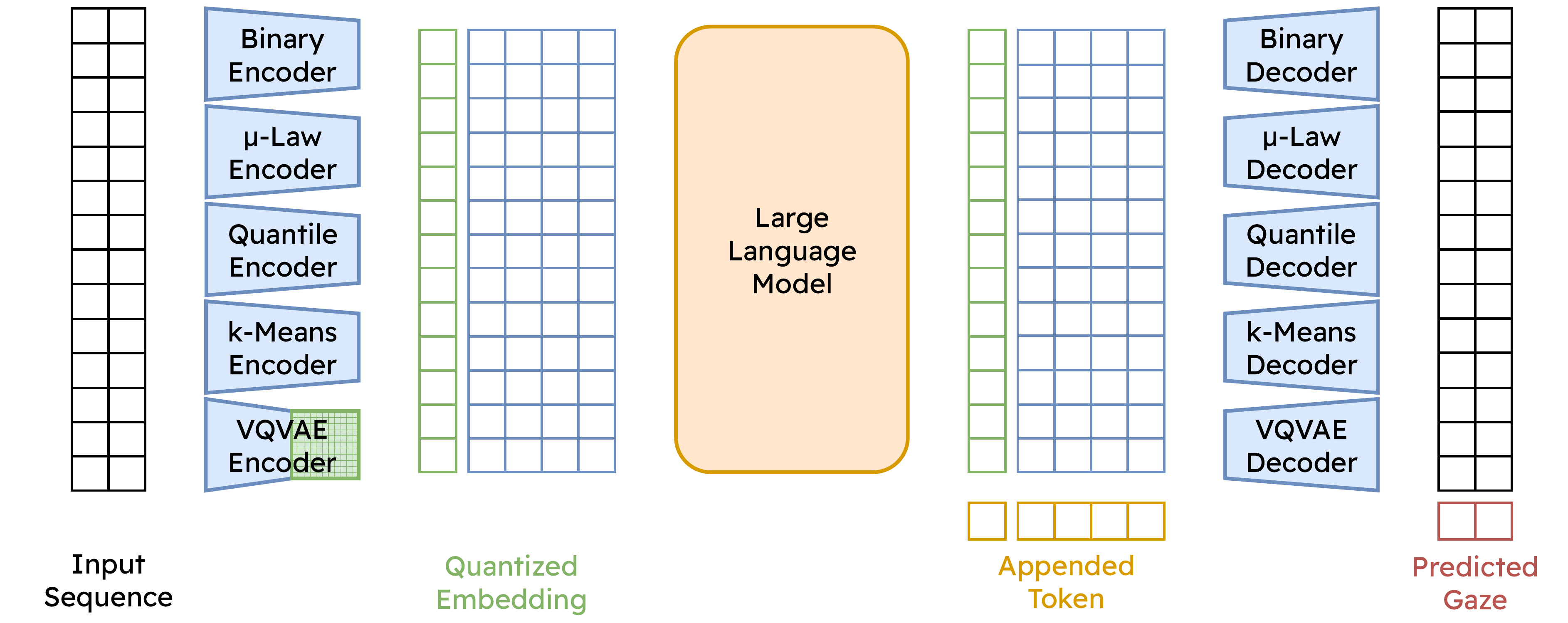}
  \caption{Different tokenization mechanisms for the forecasting or generation of gaze sequences. Note that only one encoder, decoder, and large language model combination is used at the same time.}
  \Description{Architecture overview, visualizing the steps through the encoder, codebook, large language model-based predictor and final decoder.}
  \label{fig:teaser}
\end{teaserfigure}

\received{20 February 2007}
\received[revised]{12 March 2009}
\received[accepted]{5 June 2009}

\maketitle

\section{Introduction}
Transformers, as introduced by Vaswani et al. \cite{vaswani2017attention}, are a deep-learning (DL) architecture for sequential data that have revolutionized the fields of natural language processing (NLP) and computer vision (CV).
For their input, these models commonly rely heavily on tokenization to represent input text as numerical data while simultaneously shortening input length, reducing the computation cost. Tokenization describes the process of breaking down data into smaller, finite, meaningful units called tokens. Once trained, transformers generate novel sequences by processing these tokens through stacked layers of self-attention and feed-forward networks, capturing complex dependencies and contextual information predicting the next token from previous input. In the realm of NLP, tokens often represent words or frequently occurring sequences of characters, while in CV, tokens commonly correspond to 
image patches, 
allowing various tasks such as generating novel texts~\cite{radford2019language}, embed multi-modal data~\cite{girdhar2023imagebind}, or classifying~\cite{dosovitskiy2020image} and editing~\cite{chang2023muse} images.\\

Yet, when it comes to multi-modal and especially gaze data, there is, to the best of our knowledge, no consensus on a good tokenization scheme or research on tokenization methods used for gaze data that captures the underlying structure.
Consider recent papers that utilize transformers, which perform short-term gaze forecasting in virtual reality~(VR)~\cite{rolff2022gazetransformer}, time-to-saccade prediction~\cite{rolff2023deep}, emotion recognition~\cite{wang2021emotion}, eye-movement classification \cite{fuhl2023tiny}, or locomotion prediction~\cite{bremer2024predicting}. In all papers, by not optimizing the tokenization strategy, these architectures might miss out on significant gains in model performance due to different factors, such as noise or repeated input due to long fixation durations.
Here, a well-thought-out tokenization strategy would allow representing fixations as singular tokens, enabling longer context lengths providing a richer history of previous gaze data along with the utilization of pretrained transformer models.\\

Even though there are some recent attempts for either domain specific continuous data \cite{han2023onellm,zhang2023meta} or foundational models for time series data \cite{das2023decoder}, we believe that a more in-depth analysis of gaze data is warranted. As gaze data captures the movement of the human eye, it is a rich source of information about the user's behavior and attention \cite{kroger2020does} to be utilized in different domains such as gaze forecasting~\cite{rolff2022gazetransformer,hu2021fixationnet}, gaze prediction \cite{hu2020dgaze}, time-to-saccade prediction \cite{rolff2022saccades}, scan path prediction \cite{yang2020predicting}, or the previously mentioned papers that utilize transformers. However, the lack of methods for tokenizing gaze data makes it difficult to analyze, process, or fine-tune this data with existing large language models. Unlike other types of data, such as text or images, multi-modal gaze data does not have a predefined structure (i.e., 3D vectors, screen coordinates, or angular coordinates) or singular data modality, which makes it challenging to define meaningful tokens. Here, it might be especially important to capture differences between different eye-movements, such as saccades, fixations, or smooth pursuits. As gaze data is often not the only modality in the dataset, it is important to also take into account the structure of the other modalities. These modalities could, for example, be the IMU information of a head-mounted display (HMD)~\cite{hu2020dgaze,hu2021fixationnet}, or the captured audio data from a wearable eye-tracker~\cite{grauman2023ego,grauman2024ego}.
Further, a well performing tokenization method is especially crucial when adapting pre-trained large language models (LLMs) to novel data types like gaze data because they're pre-trained on tokens.
Here, recent research has highlighted that fine-tuning is considered beneficial for LLMs on specific datasets over training from scratch \cite{hu2021lora,hu2023llm}, which may also apply to novel modalities \cite{han2023onellm}.
This method often yields superior results compared to training entirely new models from scratch. Furthermore, fine-tuned LLMs have demonstrated remarkable abilities in tasks they weren’t explicitly trained for, a capability known as zero-shot prediction \cite{kojima2022large,caron2021emerging,gruver2024large,radford2019language}.\\

\noindent To overcome the research gap of evaluating different tokenizers for gaze data, we propose and analyze different approaches for tokenization of multi-modal gaze data, with a special focus on egocentric gaze data. 
In total, we focus on five approaches, each tokenizing gaze into IDs for utilization with LLMs.\\

\noindent In summary, we provide the following contributions:
\begin{itemize}
    \item We analyze five different tokenizers for continuous data, including: Binary, $\mu$-Law, Quantile, k-Means, and vector quantized variational auto encoder (VQ-VAE).  
    \item We evaluate these tokenizers on three datasets, analyzing the output on gaze positions and gaze velocities regarding different metrics, such as compression ratio, reconstruction error, and a forecasting error with GPT-2. 
    \item We provide a fast framework for tokenization of continuous data, written in Rust and Python, that can be used to train custom tokenizers on other datasets.
\end{itemize}

\section{Related Work}
\label{sec: RelatedWork}
\subsection{Tokenization \& Large Language Models}
In language processing, tokenization describes the process of mapping characters, words or sub-words into indices. This is tremendously useful for LLM's, as they map these tokens into an embedding space \cite{vaswani2017attention}.
One of the earliest methods for tokenization was often the mapping of full whitespace separated words onto an index \cite{winograd1971procedures}. However, these require huge codebook sizes for every variation of a word. Hence, recently, with the advent of LLMs, more sophisticated algorithms have been proposed that are able to deal with unseen data. This has been shown to work successfully for NLP challenges \cite{mikolov2013distributed}, as these embed semantic data into the space itself. \citet{mikolov2013distributed} showed that the embedding vector of a country added to the vector for “capital” will result in the vector containing the name of that capital, i.e. $\text{``\emph{Vietnam}''} + \text{``\emph{capital}''} = \text{``\emph{Hanoi}''}$. Roughly, there are three different tokenization schemes. (I) word-based, mapping each word into its own token index; (II) character-based, mapping each character into its index; and (III) sub-word-based tokenizers, mapping parts of a word into an index. Common algorithms for character-based tokenization are WordPiece~\cite{song2020fast}, byte pair encoding (BPE) \cite{sennrich2015neural}, and SentencePiece \cite{kudo2018sentencepiece}. However, these methods are mainly designed to tokenize textual data, making it challenging to process different modalities, such as images, audio, temporal, or continuous data.
In recent research, other modalities have been included, such as images \cite{liu2024visual,girdhar2023imagebind,dosovitskiy2020image}, depth \cite{girdhar2023imagebind,han2023onellm}, heatmaps \cite{girdhar2023imagebind}, audio \cite{girdhar2023imagebind,han2023onellm}, or IMU data \cite{girdhar2023imagebind,han2023onellm}. Some works have tried including multiple modalities at once, such as OneLLM \cite{han2023onellm}, Gemini \cite{team2023gemini}, GPT-4o\footnote{\url{https://openai.com/index/hello-gpt-4o/}}, or ImageBind~\cite{girdhar2023imagebind}.\\

When it comes to continuous data, there are multiple approaches to encode it into tokens. For example, GATO~\cite{reed2022generalist} uses a $\mu$-law transformation to encode continuous signals into discrete indices. In contrast, ImageBind~\cite{girdhar2023imagebind} and OneLLM~\cite{han2023onellm} utilize a convolutional layer to project IMU data into the embedding space of the transformer. Others have tried to utilize clustering methods, such as k-means or other clustering methods \cite{yang2023teal,lakhotia2021generative}. Recently, a different approach was proposed, encoding a short temporal input sequence projected through a residual network as a token~\cite{das2023decoder,nie2022time}. Another tokenization scheme are VQVAE's \cite{van2017neural} that are used in transformer-based vision models such as DALL-E \cite{openai2021dalle}, MaskGIT \cite{chang2022maskgit} and Muse \cite{chang2023muse}.
%
%
Despite those recent advancements, there is, to the best of our knowledge, no tokenization scheme specifically designed or tested with gaze in mind. However, it is possible to adapt tokenizers working with continuous data for this problem. For a detailed description of each tokenization method used in our paper, see \autoref{sec: Tokenization Methods}.

\section{Methodology}
In this work, we analyze various tokenization methods for eye movement data. First, we will introduce metrics for the evaluation of tokenization schemes. Afterward, we will introduce the tokenization methods utilized. Generally, we require a tokenization of continuous gaze data that minimizes its reconstruction error, meaning how well an encoded sequence can be decoded into its original sequence. This is required to feed accurate data into the LLM and correctly reinterpret the output of the LLM. Given the tokenization scheme, LLM embeddings should then capture semantic relationships and features of the data to produce meaningful predictions. Moreover, given a sequence with a high number of fixations, with small deviation, it may even make sense to compress all adjacent fixation samples into a singular token. Hence, the generated tokens should also be compressed into fewer tokens to provide a longer context length and thus more processed gaze points while simultaneously reducing data imbalance.
Similarly, byte pair encoding (BPE) \cite{sennrich2015neural,schuster2012japanese} has gained substantial attention in the NLP community for its ability to compress frequent token combinations into a new token, allowing to compress the input considerably \cite{goldman2024unpacking}. As a gaze sequence is made up of a considerable number of fixations, it would make sense to follow the same procedure by compressing them into a smaller number of samples. 
However, due to the small dataset sizes, we do not expect BPE models to outperform the non BPE models. 


\subsection{Metrics}
\label{sec: Metrics}
For our evaluation suite, we propose a set of different metrics that capture the performance of a tokenizer regarding gaze data. We specifically choose some metrics with gaze data in mind, while using more general metrics to assess various aspects of the analyzed tokenizers. Often, it is not unnatural to utilize evaluation methods that quantify the effectiveness of a tokenizer regarding the performance of a model \cite{goldman2024unpacking,kudo2018subword,provilkov2019bpe,vilar2021statistical}.
In total, we evaluate all metrics on gaze position and velocity, as both may be valuable for encoding gaze sequences.


\paragraph{\textbf{Reconstruction metrics:}} The reconstruction error describes how well the tokenized sequence can be reconstructed into the original signal. Following \citet{rolff2022gazetransformer}, we use the \textbf{mean-square error} (\textbf{MSE}) and \textbf{mean-absolute error}~(\textbf{MAE}) for the reconstruction error of the continuous gaze data. 
Ideally, the tokenization function $f$ should be bijective, 
allowing to compute the identity of the input $f^{-1}(f(x)) = x$. Given a perfect reconstruction function, it would result in an optimal reconstruction error of zero (cf. binary tokenizer of \cref{tab: reconstruction} and  \cref{tab: features} in the appendix).

\paragraph{\textbf{Compression metrics:}}
To assess the effectiveness of BPE, we measure the compression ratio \mbox{($= \frac{\text{uncompressed length}}{\text{compressed length}}$)} and space-saving \mbox{($= 1 - \frac{\text{compressed length}}{\text{uncompressed length}}$)} comparing against the uncompressed byte sequence \cite{aruna2023analysis}.

\paragraph{\textbf{Model evaluation:}}
Besides the evaluation of the reconstruction and compression parameters, it is also helpful to understand how well a model performs given a tokenizer. In our case, we use the well established pre-trained GPT-2 \cite{radford2019language} transformer architecture that allows us to fine-tune to our new datasets. Here, we will evaluate each model regarding its performance on forecasting and sequence generation using MSE and MAE metrics to evaluate the forecasting. Furthermore, we utilize \textbf{dynamic-time warping}~(\textbf{DTW})~\cite{sakoe1971dynamic,sakoe1978dynamic,cuturi2017soft}, to allow the evaluation of time shifted sequences. The reasoning behind this is the observation that the signal might be reconstructed one sample later. This may result in a high \emph{MSE} in the case of a saccade, even though it might have faithfully reconstructed the sequence, just slightly shifted. To evaluate the generation performance of the models, we use the \textbf{Jenson Shannon Divergence} (\textbf{JSD}) \cite{prasse2023sp} on position and velocity. For evaluation, we utilize a $128\times 128$ histogram generated from the gaze positions and gaze velocities. We min-max normalize the data based on the ground truth and discard any invalid positions generated. Hence, if the generated data does not fall into the range of the histogram, they will not be utilized for evaluation.

\subsection{Tokenization Methods}
\label{sec: Tokenization Methods}
In the following section, we will introduce all tokenization methods used to tokenize the data and to train our models. We analyzed commonly utilized tokenizers for other applications that require continuous data. More detailed information on all tokenization methods can be found in the appendix (Sec. \ref{sec: binary tokenizer} to Sec. \ref{sec: vqvae tokenizer}). Note that we do not utilize forecasting models, such as TimesFM \cite{das2023decoder}, as they are not designed for data generation.

\paragraph*{\textbf{Binary tokenization}\nopunct} describes the process of interpreting tokens directly from the binary data, reinterpreting the float values as bytes, mapping them into an index space of 256 discrete values. Hence, a singular 4 byte gaze value results in 4 tokens with indices from 0 to 255. The tokenized length of the data will thereby increase 4-fold. 

\paragraph*{\textbf{Quantization tokenization}\nopunct} utilizes a binning function to group similar values together and separate significantly differing ones. One example of such a tokenization method is the $\mu$-law transformation, used by GATO~\cite{reed2022generalist} and WaveNet~\cite{van2016wavenet}.
Designing an optimal transformation function for binning is a non-trivial problem. Hence, we follow previous literature and use a \textbf{$\mu$-law} transformation for our data. However, as this function is not always the most optimal, we also provide another tokenizer that uses \textbf{quantile binning} to evenly quantize the data, such that each bin has roughly the same number of samples. However, it would also be possible to use any other transformation function instead. In contrast to binary tokenization, these methods allow selecting the desired number of tokens for a tradeoff between accuracy and occupied embedding vectors. For both tokenizers, we use a vocabulary size of 2048 and optimized the $\mu$-law parameters through Bayesian hyperparameter optimization \cite{snoek2012practical}, which can be found in \cref{sec: mu-law tokenizer}.

\paragraph*{\textbf{K-Means tokenization}\nopunct} describes the utilization of the cluster indices generated through k-means clustering \cite{macqueen1967some} as tokens \cite{yang2023teal}.
This allows to compress both horizontal and vertical coordinates into a single index. To train our k-means clustering, we used a convergence factor of $0.0001$ and stopped at $300$ iterations. 
We initialized the k-means clusters with the k-means++ algorithm \cite{arthur2006k}. Analyzing the initial vocabulary size (cf. appendix~\cref{fig: optimal token count}) to reduce the reconstruction error with k-means, we found a token count of $2048$ tokens to be optimal regarding performance and training tradeoff.

\paragraph*{\textbf{Vector Quantized Variational Autoencoder (VQ-VAE) \cite{van2017neural} tokenization}\nopunct} builds on the idea of variational auto encoders \cite{kingma2013auto}. This allows models to learn to embed and reconstruct data into and from a latent space. To prevent posterior collapse \cite{van2017neural}, VQ-VAE's quantize the data into a finite codebook of vectors, conveniently providing a token index. Through choice of architecture, it is possible to utilize context information for the construction of the embedding space.
Similarly to K-Means tokenization, the VQ-VAE automatically models the data distribution with the added benefit of the semantics not being restricted to the original feature space. Instead, the VQ-VAE projects the data into its latent feature space, possibly taking information from the previous gaze samples. To avoid temporal data leakage and for feature extraction, our architecture utilizes causal convolution layers \cite{van2016wavenet}, only processing information from the current and previous data samples. As the last layer, we use an LSTM \cite{hochreiter1997long} to capture long-range characteristics and dependencies of the sequence. The decoder mirrors the encoder's architecture, with the causal convolution layers being replaced by transposed convolutions. We again set our initial vocabulary / codebook size to 2048 entries and employ a strategy to explicitly replace rarely used codewords with embeddings that are not well represented in the codebook, following the replacement strategy of \citet{dhariwal2020jukebox} and \citet{huh2023straightening}. Equally to \citet{lancucki2020robust}, we initialize the codebook through k-means++\cite{arthur2006k}, to ensure an immediately useful codebook and correct scaling of the quantized embeddings. The model is trained using MSE loss for 35000 epochs, as we found it to still converge at high epochs. The model architecture and training parameters can be found in \cref{sec: vqvae tokenizer}.

\section{Results}
We evaluate the selected tokenizers on three datasets, namely: Ego-Exo4D \cite{grauman2024ego}, DGaze \cite{hu2020dgaze}, and FixationNet \cite{hu2021fixationnet}. In total, all datasets contain 12,224,614 (Ego-Exo4D), 1,840,435 (dgaze), and 2,077,054 (FixationNet) gaze samples. All of these were captured in egocentric vision either in VR (Dgaze, FixationNet) or real-world (Ego-Exo4D) with DGaze and FixationNet capturing samples at 100Hz and Ego-Exo4D at 10Hz.
\paragraph{\textbf{Reconstruction Error:}\nopunct}
\label{sec: reconstruction error}
\begin{table*}[t]
    \centering
    \caption{Reconstruction results of our analyzed tokenizers on the metrics listed in~\cref{sec: Metrics}. Note that we do not have an accumulative error for positional data. For an explanation, see \cref{sec: reconstruction error}. A lower error on all metrics is preferred. Except binary, the best tokenizer are marked in \best{bold}.}
    \resizebox{0.9\textwidth}{!}{%
            \begin{tabular}{cc|c|c|c|c|c|c|c|c|c|c|c}
       \toprule
         \rowcolor{LightGreen!40}
       & & 
       & \multicolumn{5}{c|}{Individual Error}
       & \multicolumn{5}{c}{Accumulative Error}\\
       \cline{4-8}
       \cline{9-13}
       \rowcolor{LightGreen!40}
       \multirow{-2}{*}{Distribution}
       & \multirow{-2}{*}{Dataset}
       & \multirow{-2}{*}{Metric}
       & \multicolumn{1}{c|}{Binary}
       & \multicolumn{1}{c|}{$\mu$-Law}
       & \multicolumn{1}{c|}{Quantile}
       & \multicolumn{1}{c|}{k-Means}
       & \multicolumn{1}{c|}{VQ-VAE}
       & \multicolumn{1}{c|}{Binary}
       & \multicolumn{1}{c|}{$\mu$-Law}
       & \multicolumn{1}{c|}{Quantile}
       & \multicolumn{1}{c|}{k-Means}
       & \multicolumn{1}{c}{VQ-VAE}\\
       \midrule
        \rowcolor{LightCyan}
        & 
         & MSE$_\downarrow$
          & 0.000$^\circ$ & 0.024$^\circ$ & 0.019$^\circ$ & 0.110$^\circ$ & \best{0.000$^\circ$}
          & --- & --- & --- & --- & ---\\
        \rowcolor{LightCyan}
        & \multirow{-2}{*}{Ego-Exo4D} 
         & MAE$_\downarrow$
          & 0.000$^\circ$ & 0.133$^\circ$ & 0.013$^\circ$ & 0.257$^\circ$ & \best{0.013$^\circ$}
          & --- & --- & --- & --- & ---\\
       \cline{1-9}
       \rowcolor{LightCyan!50}
       & 
        & MSE$_\downarrow$ 
         & 0.000$^\circ$ & 0.062$^\circ$ & 0.006$^\circ$ & 0.100$^\circ$ & \best{0.000$^\circ$}
         & --- & --- & --- & --- & ---\\
       \rowcolor{LightCyan!50}
       & \multirow{-2}{*}{DGaze}
        & MAE$_\downarrow$ 
         & 0.000$^\circ$ & 0.216$^\circ$ & 0.014$^\circ$ & 0.238$^\circ$ & \best{0.012$^\circ$}
         & --- & --- & --- & --- & ---\\
       \cline{1-9}
       \rowcolor{LightCyan}
       &  
        & MSE$_\downarrow$ 
         & 0.000$^\circ$ & 0.051$^\circ$ & 0.011$^\circ$ & 0.096$^\circ$ & \best{0.001$^\circ$}
         & --- & --- & --- & --- & ---\\
       \rowcolor{LightCyan}
       \multirow{-6}{*}{Position}
       &  \multirow{-2}{*}{FixationNet}
        & MAE$_\downarrow$ 
         & 0.000$^\circ$ & 0.194$^\circ$ & 0.056$^\circ$ & 0.210$^\circ$ & \best{0.013$^\circ$}
         & --- & --- & --- & --- & ---\\

       \midrule

        \rowcolor{LightRed}
        & 
         & MSE$_\downarrow$ 
          & 0.000$^\circ$ & \best{0.002$^\circ$} &  8.571$^\circ$ & 0.423$^\circ$ & 0.056$^\circ$
          & 0.000$^\circ$ & 80.507$^\circ$ & 36.571$^\circ$ & 6.288$^\circ$ & \best{0.178$^\circ$}\\
        \rowcolor{LightRed}
        & \multirow{-2}{*}{Ego-Exo4D} 
         & MAE$_\downarrow$ 
          & 0.000$^\circ$ & \best{0.043$^\circ$} & 0.635$^\circ$ & 0.472$^\circ$ & 0.074$^\circ$
          & 0.000$^\circ$ & 7.760$^\circ$ & 1.995$^\circ$ & 1.716$^\circ$ & \best{0.310$^\circ$}\\
        \rowcolor{LightRed}
       \cline{1-9}
       \rowcolor{LightRed!50}
       & 
        & MSE$_\downarrow$ 
         & 0.000$^\circ$ & \best{0.005$^\circ$} &  3.452$^\circ$ & 0.206$^\circ$ & \best{0.005$^\circ$}
         & 0.000$^\circ$ & 236.087$^\circ$ & 79.601$^\circ$ & 2.487$^\circ$ & \best{0.094$^\circ$}\\
       \rowcolor{LightRed!50}
       & \multirow{-2}{*}{DGaze}
        & MAE$_\downarrow$ 
         & 0.000$^\circ$ & 0.062$^\circ$ & 0.333$^\circ$ & 0.355$^\circ$ & \best{0.027$^\circ$}
         & 0.000$^\circ$ & 13.260$^\circ$ & 3.920$^\circ$ & 1.084$^\circ$ & \best{0.224$^\circ$}\\
       \cline{1-9}
       \rowcolor{LightRed}
       & 
        & MSE$_\downarrow$ 
         & 0.000$^\circ$ & \best{0.004$^\circ$} &   4.328$^\circ$ & 0.222$^\circ$ & 0.056$^\circ$
         & 0.000$^\circ$ & 25.264$^\circ$ & 167.613$^\circ$ & 1.053$^\circ$ & \best{0.178$^\circ$}\\
       \rowcolor{LightRed}
       \multirow{-6}{*}{Velocity}
       & \multirow{-2}{*}{FixationNet}
        & MAE$_\downarrow$ 
         & 0.000$^\circ$ & \best{0.051$^\circ$} &  0.613$^\circ$ & 0.365$^\circ$ & 0.074$^\circ$
         & 0.000$^\circ$ & 4.346$^\circ$ & 10.562$^\circ$ & 0.606$^\circ$ & \best{0.310$^\circ$}\\
       \rowcolor{LightRed}
       \bottomrule
    \end{tabular}%
    }
    \label{tab: reconstruction}
\end{table*}
Given the reconstruction results shown in \cref{tab: reconstruction}, it is not unexpected that a binary tokenizer performs best, as it just reinterprets the data without applying a transformation. Further, we found a low reconstruction error for the VQ-VAE for gaze positions, which was surprisingly outperformed by the $\mu$-law tokenizer on gaze velocities. Besides the direct comparison between individual samples, we also measure the accumulative error $E_{a}$, essentially reconstructing the $i^\text{th}$ position $p_i$ from the previous velocities $v$ through $p_i = p_0 + \sum_{j=1}^i v_j$ measuring the accumulative error between the ground truth $\hat{p}$ as $E_a = \mathcal{D}(\hat{p}, p)$, with $\mathcal{D}$ being either the MAE, or MSE.
We found that a faithful reconstruction of the positions from the velocity is not viable with some selected tokenizers due to the accumulated error over each step between the tokenized velocity vectors and the ground truth data (cf.~\cref{tab: reconstruction}). This is especially apparent for the $\mu$-law tokenizer, as it cannot reconstruct the sequence at all due to the data distribution not following the transformation function. This phenomenon can also be found with the quantile tokenizer when reconstructing the gaze velocities, even though the accumlative error of the quantile tokenizer is not as high as the $\mu$-law tokenizer. 


\paragraph{\textbf{Compression:}\nopunct}
\begin{table*}[t]
    \centering
    \caption{Compression ratios of our analyzed tokenizers on all three datasets. As basis, we compare against the binary representation of the data. A higher value on all metrics is preferred. Best results are marked in \best{bold}.}
    \resizebox{0.9\textwidth}{!}{%
            \begin{tabular}{c|c|c|c|c|c|c|c|c|c|c|c|c}
       \toprule
         \rowcolor{LightGreen!30}
       & & 
       & \multicolumn{5}{c|}{Non Compressed}
       & \multicolumn{5}{c}{Compressed (BPE)}\\
       \cline{4-8}
       \cline{9-13}
       \rowcolor{LightGreen!30}
       \multirow{-2}{*}{Distribution} & \multirow{-2}{*}{Dataset} & \multirow{-2}{*}{Metric} & Binary & $\mu$-Law & Quantile & k-Means & VQ-VAE & Binary & $\mu$-Law & Quantile & k-Means & VQ-VAE \\
       \midrule
        \rowcolor{LightCyan}
        & 
         & Ratio$_\uparrow$ 
          & 1.00 & 4.00 & 4.00 &  \best{8.00} & 4.00
          & 1.92 & 8.59 & 4.19 & \best{13.72} & 4.85\\
        \rowcolor{LightCyan}
        & \multirow{-2}{*}{Ego-Exo4D} 
         & Percent$_\uparrow$
          &  0.00\% & 75.00\% & 75.00\% & \best{87.50\%} & 75.00\%
          & 47.71\% & 88.29\% & 76.11\% & \best{92.62\%} & 79.26\% \\
       \cline{2-9}
       \rowcolor{LightCyan!50}
       & 
        & Ratio$_\uparrow$ 
         & 1.00 &  4.00 & 4.00 &  \best{8.00} & 4.00
         & 3.97 &  4.01 & 4.06 & \best{39.48} & 4.25\\
       \rowcolor{LightCyan!50}
       & \multirow{-2}{*}{DGaze}
        & Percent$_\uparrow$
         &  0.00\% & 75.00\% & 75.00\% & \best{87.50\%} & 75.00\%
         & 74.79\% & 88.29\% & 75.34\% & \best{97.45\%} & 76.44\% \\
       \cline{2-9}
       \rowcolor{LightCyan}
       &  
        & Ratio$_\uparrow$ 
         & 1.00 &  4.00 & 4.00 &  \best{8.00} & 4.00
         & 7.16 &  4.00 & 7.17 & \best{44.77} & 5.37\\
       \rowcolor{LightCyan}
       \multirow{-6}{*}{Position}
       & \multirow{-2}{*}{FixationNet}
        & Percent$_\uparrow$
         &  0.00\% & 75.00\% & 75.00\% & \best{87.50\%} & 75.00\%
         & 85.89\% & 75.01\% & 85.92\% & \best{97.71\%} & 81.33\%\\

       \midrule

        \rowcolor{LightRed}
        & 
         & Ratio$_\uparrow$ 
          & 1.00 & 4.00 & 4.00 &  \best{8.00} & 4.00
          & 2.30 & 7.61 & 4.10 & \best{11.19} & 6.16\\
        \rowcolor{LightRed}
        & \multirow{-2}{*}{Ego-Exo4D} 
         & Percent$_\uparrow$
          &  0.00\% & 75.00\% & 75.00\% & \best{87.50\%} & 75.00\%
          & 53.24\% & 86.76\% & 75.60\% & \best{91.00\%} & 83.65\%\\
       \cline{2-9}
       \rowcolor{LightRed!50}
       & 
        & Ratio$_\uparrow$ 
         &  1.00 & 4.00 & 4.00 &  \best{8.00} & 4.00
         & 10.35 & 6.41 & 5.68 & \best{25.53} & 10.12\\
       \rowcolor{LightRed!50}
       & \multirow{-2}{*}{DGaze}
        & Percent$_\uparrow$
         &  0.00\% & 75.00\% & 75.00\% & \best{87.50\%} & 75.00\%
         & 90.29\% & 84.40\% & 82.37\% & \best{96.06\%} & 90.09\%\\
       \cline{2-9}
       \rowcolor{LightRed}
       & 
        & Ratio$_\uparrow$ 
         &  1.00 & 4.00 & 4.00 &  \best{8.00} & 4.00
         & 25.82 & 7.53 & 7.14 & \best{29.58}  & 22.90\\
       \rowcolor{LightRed}
       \multirow{-6}{*}{Velocity}
       & \multirow{-2}{*}{FixationNet}
        & Percent$_\uparrow$
         &  0.00\% & 75.00\% & 75.00\% & \best{87.50\%} & 75.00\%
         & 96.01\% & 86.64\% & 85.98\% & \best{96.54\%} & 95.48\%\\
       \bottomrule
    \end{tabular}%
    }
    \label{tab: compression}
\end{table*}
\Cref{tab: compression} shows the results of our compression experiment against our baseline of binary data. We found that regardless of data or compression scheme used, the k-means tokenizer performs the best. Depending on the data and tokenizer, we found marginal compression. This is especially apparent for the $\mu$-law, k-means, and VQ-VAE tokenizers, possibly due to the small dataset sizes that do not include the patterns of the test set.

\paragraph{\textbf{Forecasting:}\nopunct}
\label{sec: forecasting}
\begin{table*}[t]
    \centering
    \caption{Predictive error of GPT-2 for gaze forecasted 100ms into the future with the respective tokenizer. A lower error is preferred on all metrics. The best results are marked in \best{bold}.}
    \resizebox{0.9\textwidth}{!}{%
            \begin{tabular}{c|c|c|c|c|c|c|c|c|c|c|c|c}
       \toprule
         \rowcolor{LightGreen!30}
       & & 
       & \multicolumn{5}{c|}{Non Compressed}
       & \multicolumn{5}{c}{Compressed (BPE)}\\
       \cline{4-8}
       \cline{9-13}
       \rowcolor{LightGreen!30}
       \multirow{-2}{*}{Distribution} & \multirow{-2}{*}{Dataset} & \multirow{-2}{*}{Metric} & Binary & $\mu$-Law & Quantile & k-Means & VQ-VAE & Binary & $\mu$-Law & Quantile & k-Means & VQ-VAE \\
       \midrule
        \rowcolor{LightCyan}
        & 
         & MSE$_\downarrow$ 
          & NaN & 28.444$^\circ$ & \best{21.978$^\circ$} & 23.121$^\circ$ & 26.595$^\circ$
          & NaN & 869.009$^\circ$ & \best{32.897$^\circ$} & 120.564$^\circ$ & 43.382$^\circ$\\
        \rowcolor{LightCyan}
        & 
         & MAE$_\downarrow$
          & NaN & 3.134$^\circ$ & \best{2.346$^\circ$} & 2.645$^\circ$ & 2.720$^\circ$
          & NaN & 24.884$^\circ$ & \best{2.979$^\circ$} & 7.299$^\circ$ & 4.089$^\circ$\\
        \rowcolor{LightCyan}
        & \multirow{-3}{*}{Ego-Exo4D} 
         & DTW$_\downarrow$ 
          & NaN &  5.000 & \best{3.744} &  4.199 & 4.359
          & NaN & 38.193 & \best{4.741} & 11.501 & 6.549\\
       \cline{2-9}
       \rowcolor{LightCyan!50}
       & 
        & MSE$_\downarrow$ 
         & NaN & 3483363.068$^\circ$ & 108.993$^\circ$ & 66.796$^\circ$ & \best{59.587}$^\circ$
         & NaN & 2700240.070$^\circ$ & 55.395$^\circ$ & 160.427$^\circ$ & \best{54.440$^\circ$}\\
       \rowcolor{LightCyan!50}
       & 
        & MAE$_\downarrow$
         & NaN & 1324.815$^\circ$ & 4.366$^\circ$ & 9.474$^\circ$ & \best{3.703$^\circ$}
         & NaN & 1271.054$^\circ$ & \best{3.272$^\circ$} & 7.594$^\circ$ & 3.454$^\circ$\\
        \rowcolor{LightCyan!50}
        & \multirow{-3}{*}{DGaze}
         & DTW$_\downarrow$ 
          & NaN & 7057.042 &       17.258  & 19.602 & 14.861
          & NaN & 6346.984 & \best{13.509} & 49.430 & 14.565\\
       \cline{2-9}
       \rowcolor{LightCyan}
       &  
        & MSE$_\downarrow$ 
         & NaN & 1184413.596$^\circ$ & \best{48.610$^\circ$} & 80.630$^\circ$ & 84.565$^\circ$
         & NaN & 1257333.258 & \best{59.338$^\circ$} & 114.635$^\circ$ & 68.215$^\circ$\\
       \rowcolor{LightCyan}
       \multirow{-7}{*}{Position}
       & 
        & MAE$_\downarrow$
         & NaN & 870.614$^\circ$ & \best{3.424$^\circ$} & 3.755$^\circ$ & 3.443$^\circ$
         & NaN & 934.872$^\circ$ & 4.401$^\circ$ & 7.076$^\circ$ & \best{3.023$^\circ$}\\
        \rowcolor{LightCyan}
        & \multirow{-3}{*}{FixationNet}
         & DTW$_\downarrow$ 
          & NaN & 4473.085 & \best{13.550} & 16.437 & 14.718
          & NaN & 4504.452 &       19.712  & 36.812 & \best{14.156}\\
       
       \midrule

        \rowcolor{LightRed}
        & 
         & MSE$_\downarrow$ 
          & NaN & 279.481$^\circ$ & 151.902$^\circ$ & 50.420$^\circ$ & \best{37.621$^\circ$}
          & NaN & 308.651$^\circ$ & 134.850$^\circ$ & 52.668$^\circ$ & \best{42.171$^\circ$}\\
        \rowcolor{LightRed}
        & 
         & MAE$_\downarrow$
          & NaN & 15.780$^\circ$ & 7.132$^\circ$ & 4.630$^\circ$ & \best{3.513$^\circ$}
          & NaN & 16.059$^\circ$ & 6.612$^\circ$ & 4.831$^\circ$ & \best{4.070$^\circ$}\\
        \rowcolor{LightRed}
        & \multirow{-3}{*}{Ego-Exo4D} 
         & DTW$_\downarrow$
          & NaN & 22.940$^\circ$ & 11.693$^\circ$ & 7.308$^\circ$ & \best{5.567$^\circ$}
          & NaN & 23.690$^\circ$ & 10.854$^\circ$ & 7.637$^\circ$ & \best{6.441$^\circ$}\\
       \cline{2-9}
       \rowcolor{LightRed!50}
       & 
        & MSE$_\downarrow$ 
         & NaN & 113682.853$^\circ$ & 3112.865$^\circ$ & \best{80.879$^\circ$} & 888.266$^\circ$
         & NaN & 33721.356$^\circ$ & 2083.594$^\circ$ & \best{75.244$^\circ$} & 216.733$^\circ$\\
       \rowcolor{LightRed!50}
       &
        & MAE$_\downarrow$
         & NaN & 266.008$^\circ$ & 27.278$^\circ$ & \best{4.436$^\circ$} & 8.539$^\circ$
         & NaN & 113.203$^\circ$ & 23.761$^\circ$ & \best{4.217$^\circ$} & 5.243$^\circ$\\
        \rowcolor{LightRed!50}
        & \multirow{-3}{*}{DGaze} 
         & DTW$_\downarrow$
          & NaN & 1025.408 & 97.248 & \best{19.487} & 30.269
          & NaN & 429.244 & 89.578 & \best{19.138} & 20.252\\
       \cline{2-9}
       \rowcolor{LightRed}
       & 
        & MSE$_\downarrow$ 
         & NaN & 133695.111$^\circ$ & 2859.888$^\circ$ & \best{38.996$^\circ$} & 105.263$^\circ$
         & \best{36.127$^\circ$} & 58495.939$^\circ$ & 2987.130$^\circ$ & 36.749$^\circ$ & 174.604$^\circ$\\
       \rowcolor{LightRed}
       & 
        & MAE$_\downarrow$
         & NaN & 272.710 & 35.589 & \best{2.938$^\circ$} & 5.289$^\circ$
         & \best{2.516$^\circ$} & 171.453$^\circ$ & 34.638$^\circ$ & 2.689$^\circ$ & 9.467$^\circ$\\
       \rowcolor{LightRed}
       \multirow{-9}{*}{Velocity}
       & \multirow{-3}{*}{FixationNet}
        & DTW$_\downarrow$
         & NaN & 995.502 & 28.493 & \best{12.754} & 21.159
         & \best{10.452} & 668.922 & 134.629 & 12.071 & 33.981\\
       \bottomrule
    \end{tabular}%
    }
    \label{tab: forecasting}
\end{table*}
\begin{table*}[t]
    \centering
    \caption{JSD on generated gaze sequences through GPT-2 using the respective tokenizer (cf.~\cref{sec: generation}). A lower value is preferred. The best results are marked in \best{bold}.}
    \resizebox{0.9\textwidth}{!}{%
            \begin{tabular}{c|c|c|c|c|c|c|c|c|c|c|c|c}
       \toprule
         \rowcolor{LightGreen!30}
       & & 
       & \multicolumn{5}{c|}{Non Compressed}
       & \multicolumn{5}{c}{Compressed (BPE)}\\
       \cline{4-8}
       \cline{9-13}
       \rowcolor{LightGreen!30}
       \multirow{-2}{*}{Distribution} & \multirow{-2}{*}{Dataset} & \multirow{-2}{*}{Metric} & Binary & $\mu$-Law & Quantile & k-Means & VQ-VAE & Binary & $\mu$-Law & Quantile & k-Means & VQ-VAE \\
       \midrule
        \rowcolor{LightCyan}
        & 
         & JSD$_\downarrow$ 
          & 0.394 & \best{0.209} & 0.588 & 0.359 & 0.425
          & 0.661 & 0.583 & 0.590 & \best{0.362} & 0.419\\
        \rowcolor{LightCyan}
        & \multirow{-2}{*}{Ego-Exo4D} 
         & Vel. JSD$_\downarrow$
          & 0.328 & \best{0.194} & 0.486 & 0.275 & 0.438
          & 0.441 & \best{0.239} & 0.483 & 0.274 & 0.434 \\
       \cline{2-9}
       \rowcolor{LightCyan!50}
       & 
        & JSD$_\downarrow$ 
         & 0.421 & 0.824 & \best{0.385} & 0.450 & 0.379
         & 0.692 & 0.659 & \best{0.496} & 0.640 & 0.567\\
       \rowcolor{LightCyan!50}
       & \multirow{-2}{*}{DGaze}
        & Vel. JSD$_\downarrow$
         & 0.176 & 0.626 & 0.283 & \best{0.130} & 0.119
         & 0.250 & 0.636 & \best{0.133} & 0.640 & 0.114 \\
       \cline{2-9}
       \rowcolor{LightCyan}
       &  
        & JSD$_\downarrow$ 
         & 0.451 & 0.833 & \best{0.320} & 0.341 & 0.345
         & 0.779 & 0.593 & 0.587 & \best{0.561} & 0.496\\
       \rowcolor{LightCyan}
       \multirow{-6}{*}{Position}
       & \multirow{-2}{*}{FixationNet}
        & Vel. JSD$_\downarrow$
         & 0.236 & 0.634 & \best{0.140} & 0.175 & 0.094
         & \best{0.138} & 0.623 & 0.219 & 0.641 & 0.151\\

       \midrule

        \rowcolor{LightRed}
        & 
         & Vel. JSD$_\downarrow$ 
          & \best{0.690} & 0.709 & 0.721 & 0.702 & 0.702
          & 0.738 & 0.716 & 0.720 & \best{0.690} & 0.736\\
        \rowcolor{LightRed}
        & \multirow{-2}{*}{Ego-Exo4D} 
         & Acc. JSD$_\downarrow$
          & \best{0.197} & 0.416 & 0.446 & 0.559 & 0.585
          & 0.454 & 0.468 & 0.431 & \best{0.426} & 0.719\\
       \cline{2-9}
       \rowcolor{LightRed!50}
       & 
        & Vel. JSD$_\downarrow$ 
         & 0.778 & 0.809 & 0.814 & \best{0.775} & 0.776
         & \best{0.770} & 0.782 & 0.798 & 0.787 & 0.792\\
       \rowcolor{LightRed!50}
       & \multirow{-2}{*}{DGaze}
        & Acc. JSD$_\downarrow$
         & 0.136 & 0.757 & 0.762 & \best{0.132} & 0.182
         & 0.371 & \best{0.172} & 0.736 & 0.181 & 0.685\\
       \cline{2-9}
       \rowcolor{LightRed}
       & 
        & Vel. JSD$_\downarrow$ 
         & \best{0.800} & 0.817 & 0.826 & 0.805 & 0.808
         & \best{0.801} & 0.811 & 0.815 & 0.808 & 0.821\\
       \rowcolor{LightRed}
       \multirow{-6}{*}{Velocity}
       & \multirow{-2}{*}{FixationNet}
        & Acc. JSD$_\downarrow$
         & 0.162 & 0.822 & 0.767 & \best{0.101} & 0.152
         & \best{0.123} & 0.471 & 0.771 & 0.141 & 0.242\\
       \bottomrule
    \end{tabular}%
    }
    \label{tab: generation}
\end{table*}
To evaluate the performance of our tokenizers, we use them in two tasks: (I) prediction and (II) gaze data generation. For prediction, we choose a prediction horizon of 100ms, which coincides with the latency of commercially available head-mounted display eye trackers \cite{stein2021comparison}. \Cref{tab: forecasting} shows the results when fine-tuning a GPT-2 with the tokens generated through the tokenizers. As expected, we found that the binary tokenizer performs the worst, as it simultaneously tries to learn the distribution of floating point numbers and gaze data, generally resulting in invalid numbers (NaN). Besides, we found that the $\mu$-law tokenizer performing poorly, even though its reconstruction performance was similar to the other tokenizers (cf.~\cref{tab: reconstruction}). While we do not expect the BPE results to represent a massive dataset, it still shows that the quantile and VQ-VAE tokenizers mostly outperform all others for the forecasting of gaze positions on BPE and non-BPE tokenizers. This also aligns with its good reconstructive performance shown in \cref{tab: reconstruction}. For the forecasting of gaze velocities, we found that the quantile tokenizer performs significantly worse than for the forecasting of gaze positions. Here, we observed that a k-means or VQ-VAE tokenizer performs more optimal.

\paragraph{\textbf{Generation:}\nopunct}
\label{sec: generation}
For the generation task, we choose to generate a 1 sec. sequence of data for DGaze and FixationNet and 10 sec. for the Ego-Exo4D dataset due to model restrictions. \Cref{tab: generation} shows the measured JSD on the gaze positions and gaze velocities. Furthermore, we also calculated the JSD on the second derivates of the data, meaning the velocities for the gaze positions and accelerations for the gaze velocities. This allows us to also evaluate the temporal behavior of our model. 
Remarkably, we observed non-optimal performance for the generation of gaze velocity data. While we would have expected to perform similar to \emph{Vel. JSD} of the gaze positions (cf.~\cref{tab: generation}), this is not the case. Here, further investigation is warranted, as it may be due to model choice of the LLM, inference parameters, or tokenization strategies unfit for velocity data generation. Another factor may be the computation of the histogram (cf.~\cref{sec: Metrics}), requiring a different evaluation strategy.

\section{Discussion \& Future Work}
In this paper, we evaluated five different tokenization strategies for continuous gaze data on three different datasets. We analyzed them based on their ability to accurately reconstruct the gaze signal, their compression capabilities, their forecasting error, and an analysis on how well generated data fits the test distribution. Overall, we found that a quantile or VQ-VAE tokenizer performs best on gaze positions, while a k-means tokenizer is more suited for gaze velocities. We would like to note that we just analyzed the tokenization methods and did not optimize the LLM hyperparameters or model for the task at hand. The same does apply to the LLM inference that may not be optimal for all tasks. This, however, is out of the scope of this paper and will be done in future work. It remains to be seen how much this hyperparameter optimization and model selection improves the predictive performance. It also remains to be seen if our analysis holds true for massive large-scale datasets the size of text corpora that are usually used for enormous LLMs. In conclusion, given enough data, we would recommend a VQ-VAE for the tokenization of gaze data as it provides a versatile approach that already incorporates knowledge about the data distribution. However, for smaller datasets, we would recommend a quantile tokenizer due to its ability to evenly distribute samples into bins, resulting in better training of the model. For gaze velocities, we would recommend a k-means / VQ-VAE instead of a quantile tokenizer.

\paragraph{\textbf{Potential societal impacts:}\nopunct} The prediction and analysis of one's gaze may open up some potential societal impacts. While most of them have already been discussed by \citet{kroger2020does}, for example, the prediction of someone's gender, sexual preferences, personality, or mental health, we would like to emphasize some potential impacts. As this work provides the basis for the training of LLMs for gaze data, it may be used to train a foundation model. Such a foundation model can be potentially misused for targeted advertisement or the prediction of the psychological, neurological, and medical state of a user of this technology. 
However, it can also provide a lot of potential benifits opening up thbetter utilization of gaze data in different domains. Here, we would like to advise future researchers to take these possible use cases into consideration when utilizing our work or further researching on the topic.

\paragraph{\textbf{Privacy and ethics:}\nopunct} We adhere to the ethics requirements of \emph{our institution (name removed for anonymity)}. Further, we did not capture data by ourselves or performed experiments on humans, but rather used publicly avaiable datasets. \balance 

\begin{acks}
\makeatletter
\if@ACM@anonymous
    So Long, and Thanks for All the Fish.
\else
    The research for this paper was funded by the Deutsche Forschungsgemeinschaft (DFG, German Research Foundation) -- project no. 511498220 at the University of Hamburg.
\fi
\makeatother
\end{acks}
\balance

\clearpage
\bibliographystyle{ACM-Reference-Format}
\bibliography{sample-base}

\onecolumn

\appendix
\section*{Appendix}
\section{Alternative Tokenization Schemes}
\begin{figure}[!h]
    \begin{subfigure}[t]{0.32\textwidth}
        \includegraphics[width=\textwidth]{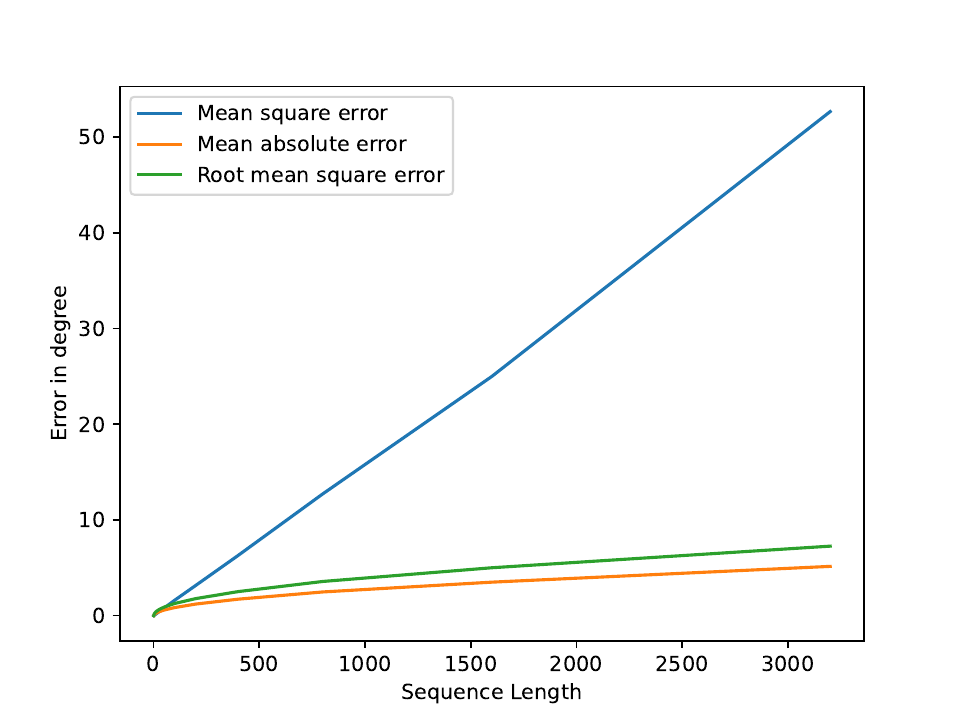}
    \end{subfigure}
    \begin{subfigure}[t]{0.32\textwidth}
        \includegraphics[width=\textwidth]{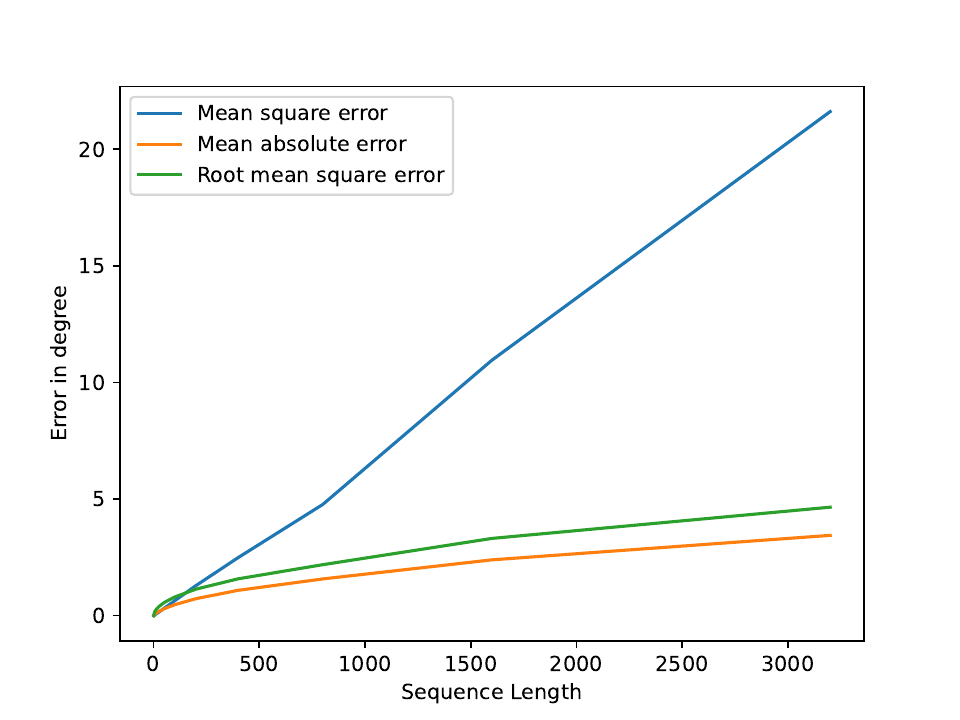}
    \end{subfigure}
    \begin{subfigure}[t]{0.32\textwidth}
        \includegraphics[width=\textwidth]{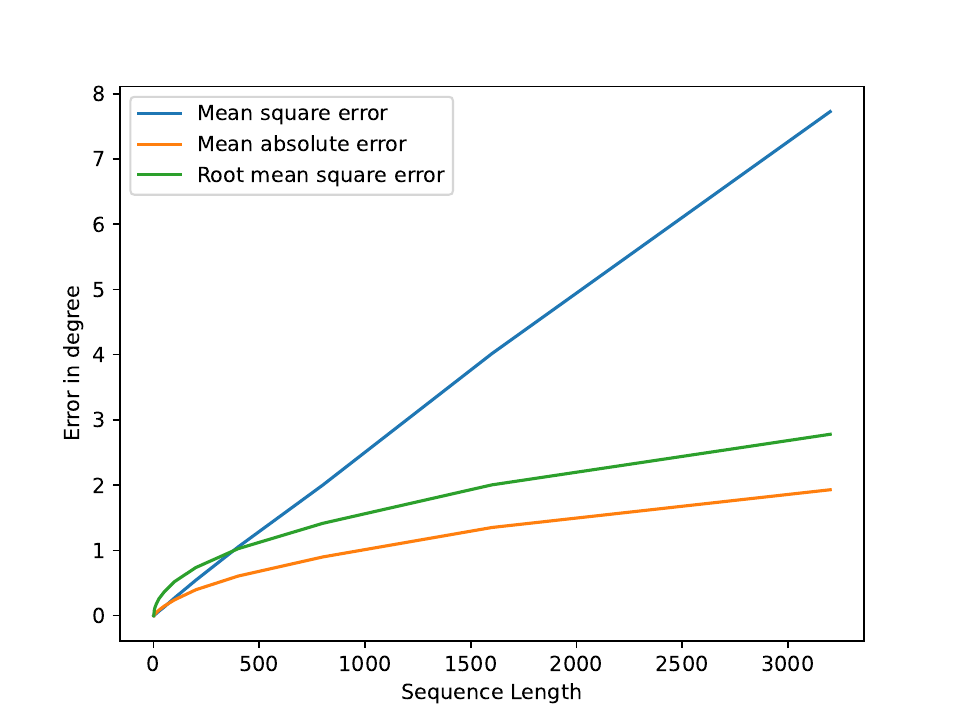}
    \end{subfigure}
    \caption{Accumulative error over the sequence length on the Ego-Exo4D (left), DGaze (middle), and FixationNet (right) datasets.}
    \label{fig: accumulative error}
\end{figure}
\noindent As shown in \cref{tab: reconstruction}, it is evident that except the VQ-VAE and the Binary tokenizers, none of the tokenization methods can satisfyingly reconstruct velocity sequences. While we assume that this is not a problem for short sequences (cf.~\cref{fig: accumulative error}) with a few samples, we expect that it might become challenging to reconstruct longer sequences accurately. Fortunately, it is possible to reduce this error accumulation, by splitting the axis of the gaze data and calculating the $\mu$-law and k-means parameter for each axis independently. 
However, this requires bookkeeping during inference of the model, as one needs to make sure to reconstruct the correct axis, especially complicated through BPE. Hence, users of either tokenizer need to make a tradeoff between the reconstruction error and a more complex inference and compression scheme.
For, a k-means tokenizer with separated axes, we were able to achieve an MSE and MAE of $0.001^\circ$ (individual) and $0.000^\circ$ (accumulative) on the Ego-Exo4D dataset. 

\begin{table}[!h]
    \centering
    \caption{Analytical differences between analyzed tokenizers.}
    \resizebox{0.7\textwidth}{!}{
    \begin{tabular}{lccccc}
         \toprule
         \textsc{Feature} &  \textsc{Binary} &  \textsc{$\mu$-Law} & \textsc{ Quantile} &  \textsc{k-Means} &  \textsc{VQ-VAE}\\
         \midrule
         Accurate reconstruction & \cmark{} & \xmark{} & \xmark{} & \xmark{} & \xmark{} \\
         Axis separation (Hor./Vert.) & \cmark{} & \cmark{} & \cmark{} & (\cmark{}) & (\cmark{})\\
         Requires training & \xmark{} & (\cmark{}) & \cmark{} & \cmark{} & \cmark{}\\
         Native compression ratio & 1:1 & 4:1 & 4:1 & 4:1 / 8:1 & ---\\
         Vocabulary size & 2048 & 2048 & 2048 & 2048 & 2048 \\
         \bottomrule
    \end{tabular}
    }
    \label{tab: features}
\end{table}

\section{Binary Tokenizer}
\label{sec: binary tokenizer}
\begin{figure}[H]
    \centering
    \includegraphics[width=0.7\textwidth,trim={0cm 10cm 12cm 0},clip]{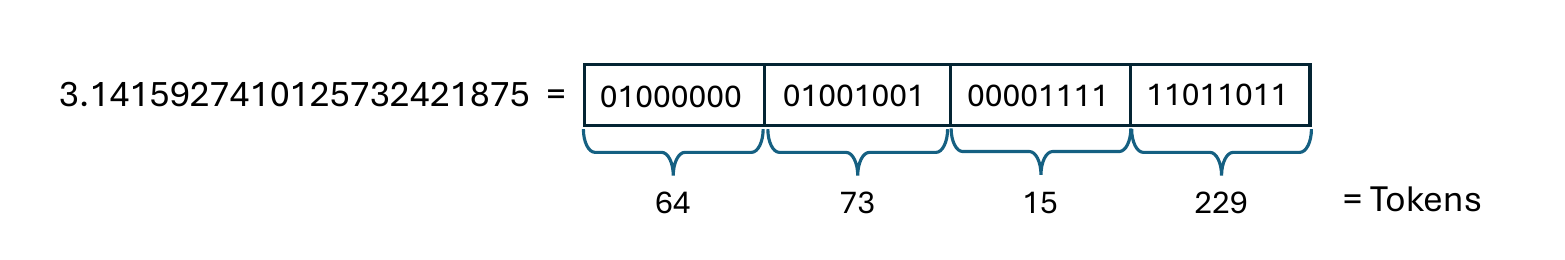}
    \caption{Examplary binary tokenization of the number $\pi$. Note that the number is not an accurate representation of $\pi$ due to IEEE float point inaccuracies.}
    \label{fig: binary tokenization}
\end{figure}
For the binary tokenization, we perform a direct reinterpret of the 32-bit IEEE float point values \cite{none2019ieee} from their binary format, as depicted in \cref{fig: binary tokenization}. Therefore, each float point value will result in four tokens. To reverse this process, we concatenate four tokens and reinterpret their byte values as a single float point value.

\section{Quantile Tokenizer}
For the quantile tokenizer, we utilize frequency binning using the quantile of the data. This results in an approximately equal distribution of tokens across all bins. Here, follow \mbox{\citet{hyndman1996sample}} for the definition of the quantile $Q(p)$ for a probabilistic data distribution $F(\text{sample})$ of our gaze samples. Therefore, $Q(p)$ is defined as:
\begin{equation*}
    Q(p) = \text{inf} \left\{ \text{sample}\ |\ F(\text{sample}) > p\right\}.
\end{equation*}
To bin the data we define a set of $n=2048$ probabilities $p_i\in[0,1]$, through: $P = \left\{p_i = \frac{i}{n}\ |\ 0\leq i < n \right\}$ and estimate the $i^{\text{th}}$ quantile as $q_i = Q(p_i)$. In practice, finding the $q_i^\text{th}$ quantile requires us to sort the list of gaze samples, returning the $\floor{p_i\cdot\text{number of samples}}^\text{th}$ element of the sorted samples.
We then utilize the found quantiles to bin the gaze samples into tokens, through:
\begin{equation*}
    \text{token index} =
    \begin{cases}
        0 & \text{sample} < q_0\\
        1 & q_0 \leq \text{sample} < q_1\\
        2 & q_1 \leq \text{sample} < q_2\\
          & \vdots \\
        n - 1 & q_{n-1} \leq \text{sample}\\
    \end{cases}
\end{equation*}\\
To reconstruct the tokenized data back into gaze samples, we again rely on the previously computed quantiles $q_i$, using the average of the two quantiles nearest to a token index, which we compute as:
\begin{equation*}
    \text{reconstructed sample} = \frac{q_{\text{token index}} + q_{\text{min}\left(\text{token index} + 1, n - 1\right)}}{2}
\end{equation*}

\section{$\mu$-Law Tokenizer}
\label{sec: mu-law tokenizer}
\begin{table}[!h]
    \centering
    \caption{Found hyperparameters for the $\mu$-law tokenizer used in our paper.}
    \resizebox{0.35\textwidth}{!}{
    \begin{tabular}{cc|c|c}
         \toprule
         Distribution & Dataset & $\mu$ & N \\
         \midrule
         \multirow{3}{*}{Position} & Ego-Exo4D & 17.324 & 0.536 \\
         & DGaze & 63.504 & 0.445\\
         & FixationNet & 35.958 & 0.422\\
         \hline
          \multirow{3}{*}{Velocity} & Ego-Exo4D & 2.000 & 1.000\\
         & DGaze & 2.000 & 1.000\\
         & FixationNet & 2.000 & 1.000\\
         \bottomrule
    \end{tabular}
    }
    \label{tab: mu-law parameters}
\end{table}
For our $\mu$-Law tokenizer, we transform the gaze data through the $\mu$-law function $f_{\mu, \text{N}}: \mathbb{R}\rightarrow[-1,1]$ as used in \cite{van2016wavenet,reed2022generalist}, defined with its inverse $f_{\mu, \text{N}}^{-1}: [-1,1]\rightarrow\mathbb{R}$ through:
\begin{equation*}
    f_{\mu, \text{N}}(x) = \text{sign}(x)\cdot \frac{\log(|x \cdot \mu| + 1)}{\log(|\mu \cdot N| + 1)},\hspace{1cm}
    f_{\mu, \text{N}}^{-1}(y) = \text{sign}(y) \cdot \frac{(1 + |\mu \cdot N|)^{|y|} - 1}{|\mu|}.
\end{equation*}
After the transformation, we bin a gaze sample into $n=2048$ bins a transformed sample into the set of tokens through $g_{\mu, \text{N}}: [-1,1]\rightarrow \text{Tokens}$ and reconstruct them using $g_{\mu, \text{N}}^{-1}:\text{Tokens}\rightarrow[-1,1]$ as follows:
\begin{equation*}
    g_{\mu, \text{N}}(\text{sample}) = \floor{n\cdot \frac{f_{\mu, \text{N}}(\text{sample}) + 1}{2}},
    \hspace{1cm}
    g_{\mu, \text{N}}^{-1}(\text{token index}) = f_{\mu,\text{N}}^{-1}\left(\frac{2\cdot \text{token index}}{n} - 1\right).
\end{equation*}
Note that $g^{-1}(g(x)) \approx x$ and is therefore not an accurate reconstruction, as we perform a floor operation during the binning. To estimate the $\mu$-Law parameters $\mu$ and $N$ we perform Bayesian hyperparameter optimization \cite{snoek2012practical}, minimizing the reconstruction error between the tokenized and reconstructed tokens as follows:
\begin{equation*}
    \underset{\mu, N}{\text{min}} \sum_{\text{gaze sample}} \left(\text{gaze sample} - g_{\mu, \text{N}}^{-1}\left(g_{\mu, \text{N}}\left(\text{gaze sample}\right)\right)\right)^2
\end{equation*}
We would like to note that we did not utilize the found parameters for gaze velocities, as they performed much worse during later stages of the valuation. Instead, we optimized these through careful exploration. The results of the hyperparameter optimization can be found in \cref{tab: mu-law parameters}. Furthermore, we also perform a min-max normalization before the $\mu$-law transformation. 
\section{K-Means Tokenizer}
\label{sec: k-means tokenizer}
\begin{figure}[H]
    \begin{subfigure}[t]{0.3\textwidth}
        \includegraphics[width=\textwidth]{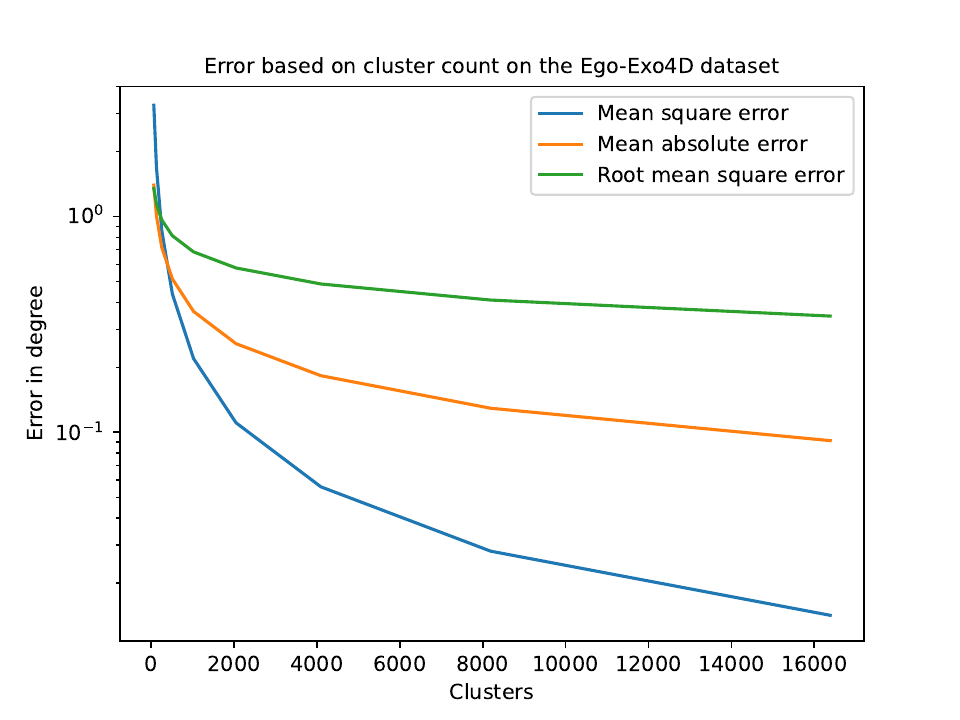}
    \end{subfigure}
    \begin{subfigure}[t]{0.3\textwidth}
        \includegraphics[width=\textwidth]{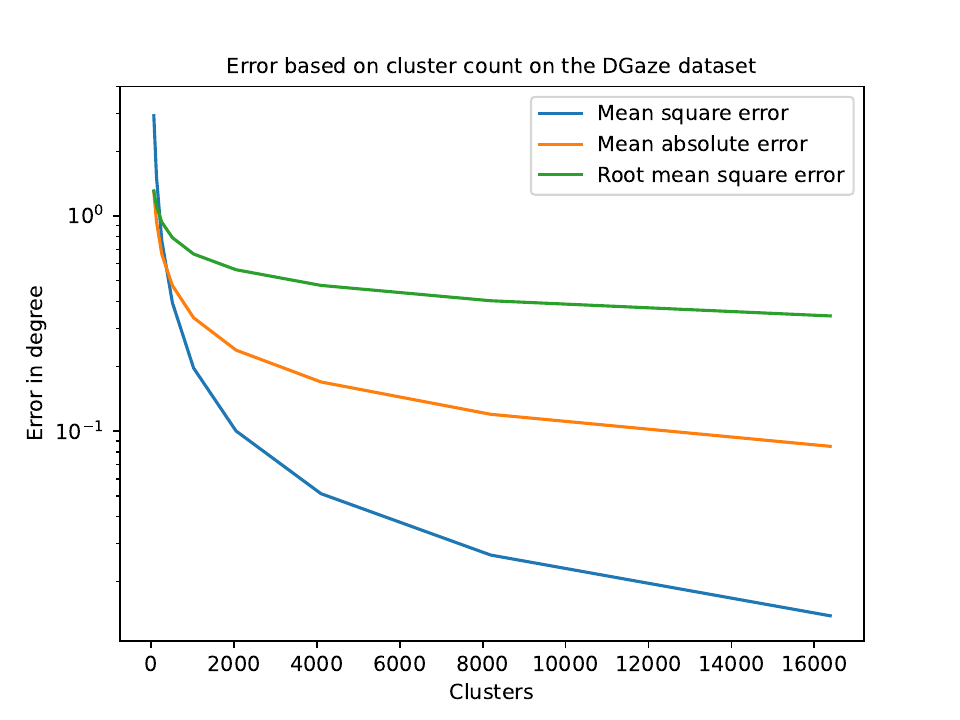}
    \end{subfigure}
    \begin{subfigure}[t]{0.3\textwidth}
        \includegraphics[width=\textwidth]{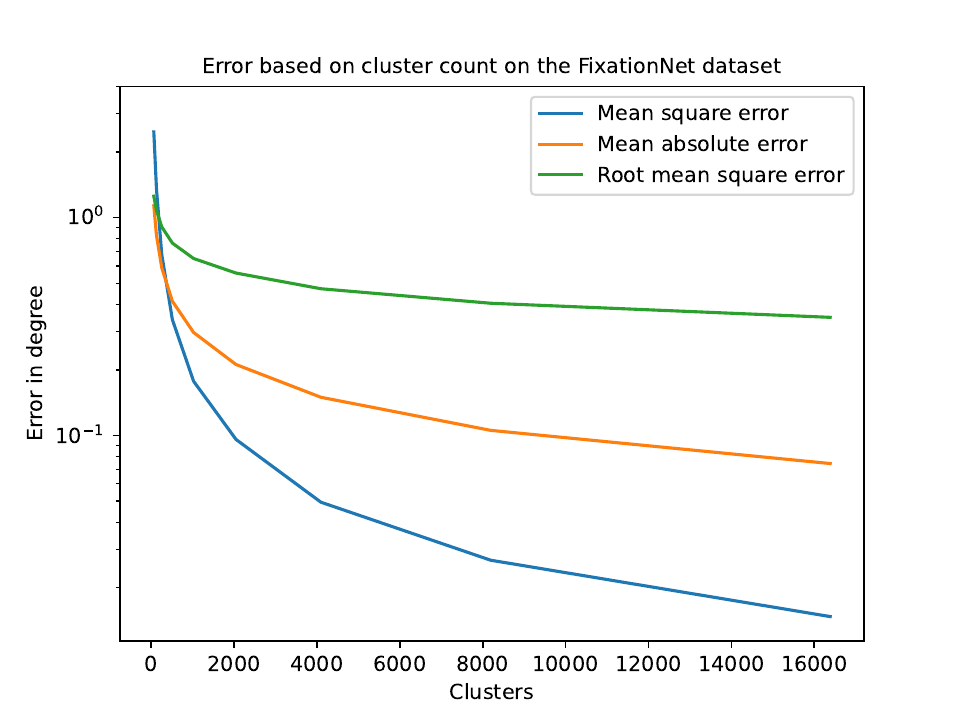}
    \end{subfigure}
    \caption{Reconstruction error of the k-means tokenizer plotted against the utilized clusters. Note: the y-axis is logarithmically scaled.}
    \label{fig: optimal token count}
\end{figure}
As mentioned in our paper, we employ k-means \cite{macqueen1967some} for finding center points that can be utilized as token indices. To compute the  $i^\text{th}$ token index for a gaze sample, we search for the nearest cluster center, $c_i$, essentially minimizing:
\begin{equation*}
    \underset{i}{\text{min}} \left\| \text{sample} - c_i \right\|_2
\end{equation*}
using the Euclidean distance $\|.\|_2$ as our distance function. To reconstruct the gaze samples, we look up the cluster center given the token index, via:
\begin{equation*}
    \text{reconstructed sample} = c_{\text{token index}}
\end{equation*}

\section{VQ-VAE Tokenizer}
\label{sec: vqvae tokenizer}
\begin{figure}[H]
    \includegraphics[width=\textwidth]{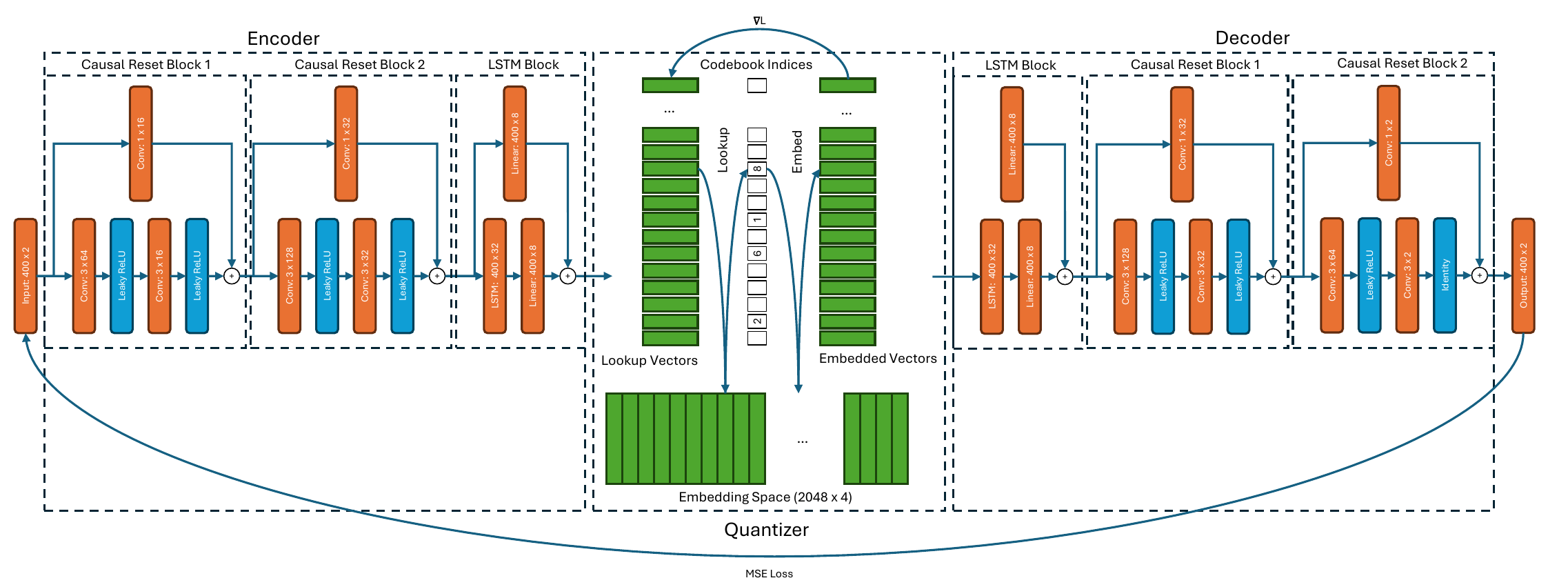}
    \caption{The VQ-VAE for the gaze position tokenizer. We utilize a codebook size of 4, resulting in two codebook entries per gaze sample. To generate the tokens, we perform inference of the encoder and use the codebook indices as our tokens.}
    \label{fig: position vqvae}
\end{figure}
\begin{figure}[H]
    \includegraphics[width=\textwidth]{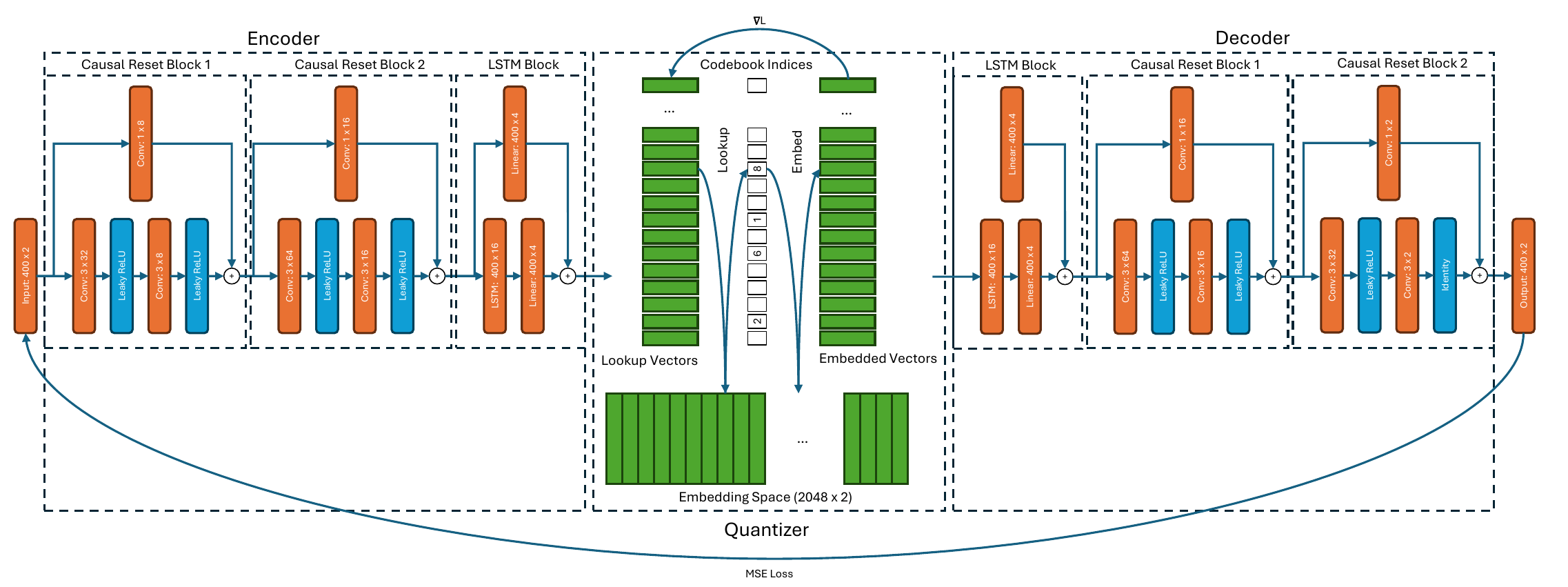}
    \caption{The VQ-VAE for the gaze velocity tokenizer. We utilize a codebook size of 2, resulting in two codebook entries per gaze sample. To generate the tokens, we perform inference of the encoder and use the codebook indices as our tokens.}
    \label{fig: velocity vqvae}
\end{figure}
\begin{figure}[H]
    \centering
    \includegraphics[width=0.85\textwidth]{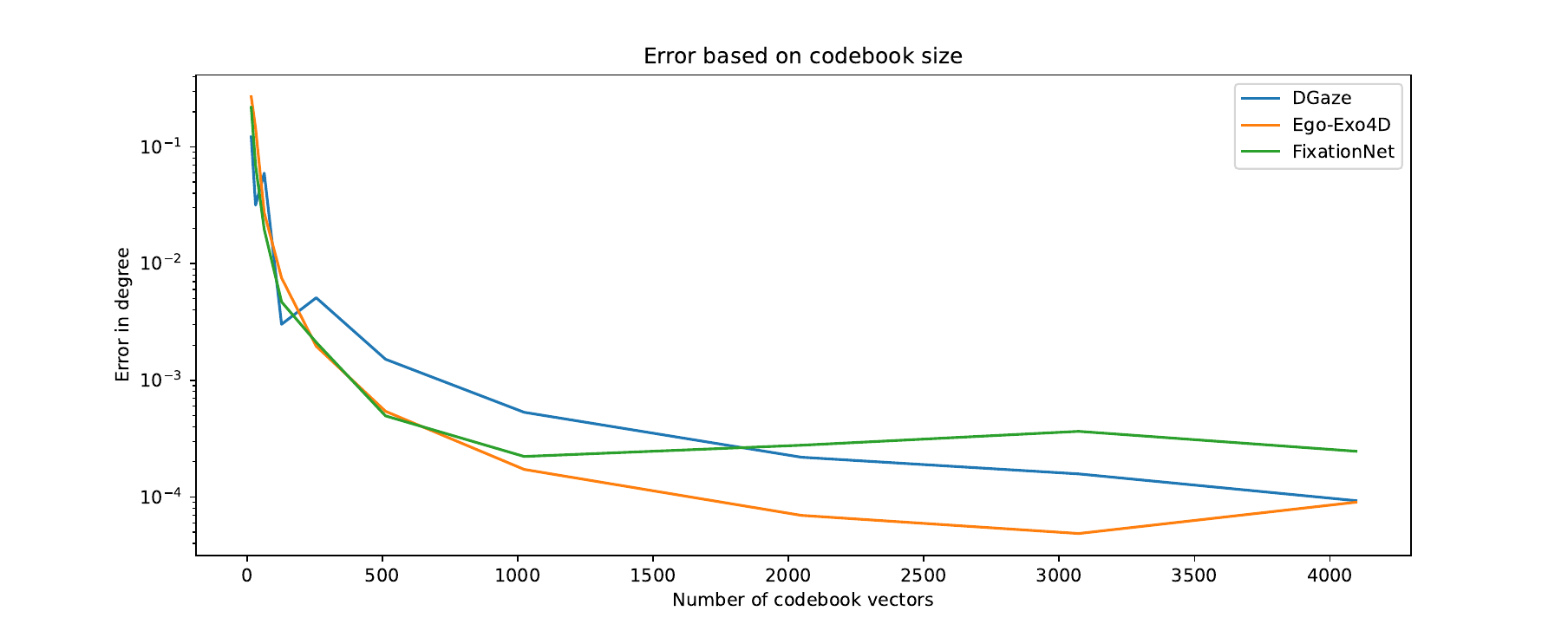}
    \caption{Measured reconstruction errors on all datasets in visual degree depending on the used codebook size of the position VQVAE.}
    \label{fig: vocabulary sizes}
\end{figure}

For the VQ-VAE tokenizer, we use two different architectures. One designed for gaze positions (cf.~\cref{fig: position vqvae}) and the other one designed for gaze velocities (cf.~\cref{fig: velocity vqvae}), with the exact layer sizes shown in \cref{tab: layer sizes}. One specialty about the utilized architecture is that we require two codebook vectors for each vector generated by the encoder. Along with the other hyperparameter choices, we empirically determined this to be the best choice in architecture, hypothesizing that each axis of a gaze sample is stored separately. To optimize the model, we employ an Adam optimizer \cite{kingma2014adam} with a learning rate of $ 0.0002$. We schedule the learning rate using a warmup of 500 epochs and anneal it to 1e-8 over the next 35000 epochs. For the codebook loss of the quantizer, we use a $\beta = 0.2$ and replace unused codebook vectors with the vector furthest in the codebook if they have not been used for the last 20 steps. We further perform normalization of the data before training. Same as to the k-means tokenizer we also analyzed different vocabulary sizes shown in \cref{fig: vocabulary sizes} with 2048 being a good tradeoff between computational requirements and reconstruction error.\\

\noindent To compute a token index, we use the encoder \emph{enc}, the quantizer \emph{quant}, and its codebook vectors $v$ as follows:
\begin{equation*}
    \text{token index} = \underset{i}{\text{min}} \left\| v_i - \text{enc}(\text{sample}) \right\|
\end{equation*}
To reconstruct a gaze sample, we use the $v_i^\text{th}$ codebook entry and perform inference on the decoder $\text{dec}$ to reconstruct the gaze sample, via:
\begin{equation*}
    \text{reconstructed sample} = \text{dec}(v_\text{token index}).
\end{equation*}

\begin{table}[t]
    \centering
    \caption{VQ-VAE architectures used in our paper. Note that the quantizer always requires two codebook vectors for one encoded vector.}
    \begin{subfigure}[t]{0.47\textwidth}
        \caption{VQ-VAE for gaze positions corresponding to \cref{fig: position vqvae}.}
        \resizebox{1\textwidth}{!}{
        \begin{tabular}{cc|c|lr}
            \toprule
            Module & Layer Name & Input Size & \multicolumn{2}{c}{Layer Parameters}\\
            \midrule
            \multirow{9}{*}{Encoder}
            & \multirow{3}{*}{Causal Resnet Block 1}
            & \multirow{3}{*}{$400\times 2$}
                & Hidden Conv & $3\times 64$\\
            & & & Output Conv & $3\times 16$\\
            & & & Residual Conv & $1\times 16$\\

            \cline{3-5}
            
            &
              \multirow{3}{*}{Causal Resnet Block 2}
            & \multirow{3}{*}{$400\times 16$}
                & Hidden Conv & $3\times 128$\\
            & & & Output Conv & $3\times 32$\\
            & & & Residual Conv & $1\times 32$\\

            \cline{3-5}
            
            &
              \multirow{3}{*}{LSTM Block}
            & \multirow{3}{*}{$400\times 32$}
                & Hidden LSTM & $32$\\
            & & & Output MLP & $8$\\
            & & & Residual MLP & $8$\\

            \hline

               \multirow{2}{*}{Quantizer}
            & 
            & \multirow{2}{*}{$400\times 8$}
                & Embedding size & 4\\
            & & & Number of embeddings & 2048\\

            \hline

               \multirow{9}{*}{Decoder}
            & \multirow{3}{*}{LSTM Block}
            & \multirow{3}{*}{$400\times 8$}
                & Hidden LSTM & $32$\\
            & & & Output MLP & $8$\\
            & & & Residual MLP & $8$\\

            \cline{3-5}
            
            &
              \multirow{3}{*}{Transposed Resnet Block 1}
            & \multirow{3}{*}{$400\times 8$}
                & Hidden Conv & $3\times 128$\\
            & & & Output Conv & $3\times 32$\\
            & & & Residual Conv & $1\times 32$\\

            \cline{3-5}
            
            &
              \multirow{3}{*}{Transposed Resnet Block 2}
            & \multirow{3}{*}{$400\times 32$}
                & Hidden Conv & $3\times 64$\\
            & & & Output Conv & $3\times 2$\\
            & & & Residual Conv & $1\times 2$\\            
            \bottomrule
        \end{tabular}        
        }
    \end{subfigure}%
    \hspace{2em}%
    \begin{subfigure}[t]{0.47\textwidth}
        \caption{VQ-VAE for gaze velocities corresponding to \cref{fig: velocity vqvae}.}
        \resizebox{1\textwidth}{!}{
        \begin{tabular}{cc|c|lr}
            \toprule
            Module & Layer Name & Input Size & \multicolumn{2}{c}{Layer Parameters}\\
            \midrule
            \multirow{9}{*}{Encoder}
            & \multirow{3}{*}{Causal Resnet Block 1}
            & \multirow{3}{*}{$400\times 2$}
                & Hidden Conv & $3\times 32$\\
            & & & Output Conv & $3\times 8$\\
            & & & Residual Conv & $1\times 8$\\

            \cline{3-5}
            
            &
              \multirow{3}{*}{Causal Resnet Block 2}
            & \multirow{3}{*}{$400\times 8$}
                & Hidden Conv & $3\times 64$\\
            & & & Output Conv & $3\times 16$\\
            & & & Residual Conv & $1\times 16$\\

            \cline{3-5}
            
            &
              \multirow{3}{*}{LSTM Block}
            & \multirow{3}{*}{$400\times 16$}
                & Hidden LSTM & $4$\\
            & & & Output MLP & $4$\\
            & & & Residual MLP & $4$\\

            \hline

               \multirow{2}{*}{Quantizer}
            & 
            & \multirow{2}{*}{$400\times 4$}
                & Embedding size & 2\\
            & & & Number of embeddings & 2048\\

            \hline

               \multirow{9}{*}{Decoder}
            & \multirow{3}{*}{LSTM Block}
            & \multirow{3}{*}{$400\times 4$}
                & Hidden LSTM & $16$\\
            & & & Output MLP & $4$\\
            & & & Residual MLP & $4$\\

            \cline{3-5}
            
            &
              \multirow{3}{*}{Transposed Resnet Block 1}
            & \multirow{3}{*}{$400\times 4$}
                & Hidden Conv & $3\times 64$\\
            & & & Output Conv & $3\times 16$\\
            & & & Residual Conv & $1\times 16$\\

            \cline{3-5}
            
            &
              \multirow{3}{*}{Transposed Resnet Block 2}
            & \multirow{3}{*}{$400\times 16$}
                & Hidden Conv & $3\times 32$\\
            & & & Output Conv & $3\times 2$\\
            & & & Residual Conv & $1\times 2$\\            
            \bottomrule
        \end{tabular}        
        }
    \end{subfigure}
    \label{tab: layer sizes}
\end{table}

\end{document}